\documentclass[letterpaper, 10 pt, conference]{ieeeconf}
\usepackage{times}
\usepackage[pdftex]{graphicx}
\usepackage{amsmath,amssymb,amsopn,amstext,amsfonts}
\usepackage{cancel}
\usepackage[space]{cite}
\usepackage{pdfsync}
\usepackage{balance}
\usepackage{color}
\usepackage{mathtools}
\usepackage{algpseudocode}
\usepackage{algorithm} 
\usepackage{bm}

\usepackage{diagbox}
\usepackage{float}
\usepackage{epstopdf}
\usepackage{pifont}
\usepackage{fixltx2e}
\usepackage{amsmath}
\usepackage{multirow}
\usepackage{url}
\usepackage{verbatim}
\usepackage[linkcolor=black,citecolor=black,urlcolor=black,colorlinks=true]{hyperref}
\usepackage{booktabs}
\usepackage{graphicx}
\usepackage{subcaption}

\algnewcommand\algorithmicforeach{\textbf{for each}}
\algdef{S}[FOR]{ForEach}[1]{\algorithmicforeach\ #1\ \algorithmicdo}

\graphicspath{{figures/}}
\DeclareGraphicsExtensions{.png,.jpg,.eps,.pdf}
\IEEEoverridecommandlockouts
\begin{document}
\title{\LARGE \bf Learning-based 3D Occupancy Prediction for Autonomous Navigation \\in Occluded Environments}
\author{Lizi Wang$^*$, Hongkai Ye$^*$, Qianhao Wang, Yuman Gao, Chao Xu and Fei Gao
\thanks{
All authors are with the State Key Laboratory of Industrial
Control Technology, Zhejiang University, Hangzhou 310027, China, and also with the Huzhou Institute of Zhejiang University, Huzhou 313000, China. $^*$ Equal contributors.
{\tt\small Email: \{lzwang, hkye, qhwangaa, ymgao, cxu and fgaoaa\}@zju.edu.cn}}}

\maketitle

\begin{abstract}
In autonomous navigation, sensors suffer from massive occlusion in cluttered environments, leaving a significant amount of space unknown.
In practice, treating the unknown space in optimistic or pessimistic ways both set limitations on planning performance. Therefore, aggressiveness and safety cannot be satisfied at the same time.
Mimicking human behavior, in this paper, we propose a method based on deep neural network to predict occupancy distribution of unknown space.
Specifically, the proposed method utilizes contextual information of environments and prior knowledge to predict obstacle distributions in the occluded space. 
Our self-supervised  learning method use unlabeled and no-ground-truth data and augments the data by simulating navigation trajectories.
Our Occupancy Prediction Network is faster than current SOTA scene completion models and is successfully applied to unseen test environments without any refinement. 
Results show that our predictor leverages the performance of a kinodynamic planner by improving security with no reduction of speed in clustered environments. 
You can find our code and video at \url{https://github.com/ZJU-FAST-Lab/OPNet}.
\end{abstract}

\section{Introduction}
\label{sec:introduction}
\begin{figure}[t]
\centering
\includegraphics[width=0.95\linewidth]{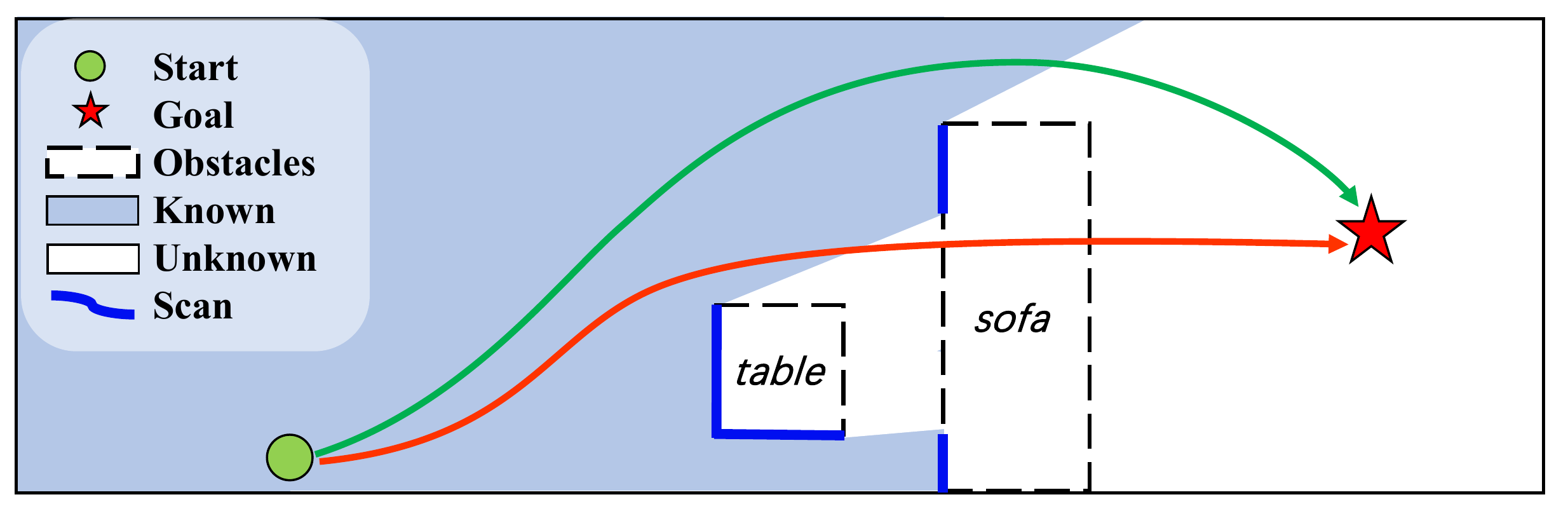}
\captionsetup{font={small}}
\caption{Planning into occluded area can cause collisions while it can be alleviated if the unrevealed occlusions can be inferred in advance.}
\label{pic:fov_occulusion}
\end{figure}

Although many works have been proposed towards autonomous navigation in unknown cluttered environments in recent years, a fast but safe scheme for light-weight platforms like UAVs and UGVs is still yet to be attained. 
Most on-board sensors, such as LiDAR and depth cameras, can only provide surface information. Therefore, only very limited surfaces of objects can be perceived as occupied while the space shaded by these surfaces remains unknown, as shown in Fig.\ref{pic:fov_occulusion}.
The unknown space, however, can be rather large in indoor environments and forests where massive occlusions are prone to happen. 
It puts the planner into a dilemma since the way to reason about the unknown space can significantly affect navigation performance.
To this end, two strategies, both with their own pros and cons, are commonly used.
One conservative manner is to treat the unknown regions as occupied and only plan in free space known to be free. It guarantees safety but limits moving speed since a stopping condition has to be met in short-range free space.
The other manner acts optimistically and treats the unknown space as free, generating aggressive trajectories into the unknown. However, it often tends to be overconfident and results in collisions.

\begin{figure}[t]
\centering
\includegraphics[width=0.95\linewidth]{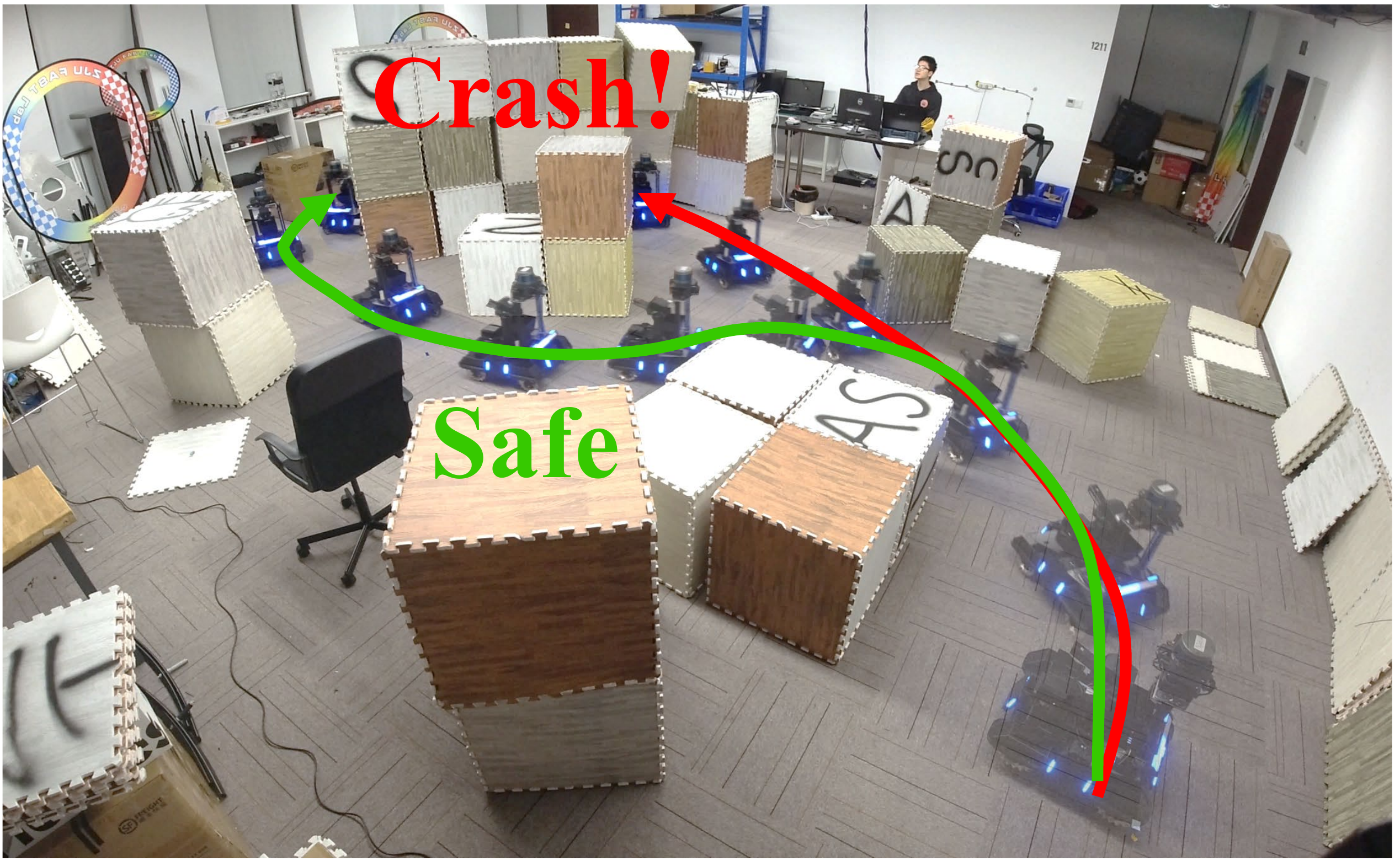}
\captionsetup{font={small}}
\caption{Composite image of UGV autonomous navigation in an unknown cluttered environment. Please watch our associated video for more demonstrations.}
\label{pic:safe_crash}
\end{figure}

In this paper, we address the above issues from the mapping perspective. 
Specifically, we opt to generate a predicted map in  the mapping module for collision check 
to achieve safe but non-conservative movements.
Firstly, we infer the occupancy distribution from partial observations of the environment and generate a more complete predicted map. 
After that, this predicted map is  used by the planner to generate smoother trajectories and avoid possible collisions in advance. Thus, the two questions we focus on are: (1) how to do occupancy prediction efficiently; (2) how to utilize occupancy prediction in our navigation system.

Data collection has long been a tiring work for deep learning tasks. In order to produce training data for occupancy prediction, we introduce a self-surpervised method  which does not require data's ground-truth or completeness. We also propose Occupancy Prediction Network (OPNet), a faster 3D fully-convolutional network with affordable computation burden. 
Furthermore, to leverage the performance of autonomous navigation, we provide a systematic solution combining basic mapping module and the proposed map predictor. Simulation and real-world navigation experiments show our system's  efficiency and robustness. Our module can be easily adjusted by a few parameters and can be applied to different map formats. 

The main contributions of this paper are:
 
\begin{enumerate}
\item A self-surpervised method for occupancy prediction. 
We augmentate training data by mimicing observation of navigation trajectories.
Learning from easily accessible data, neural networks can not only fill holes on objects, but also predict large occluded space.  

\item A lightweight yet effective 3D neural network OPNet to predict occupancy of the occluded space.
Tailored for occupancy prediction task, it is at least two times faster than current state-of-the-art scene completion models, fast enough for real-time navigation of light-weight UAVs and UGVs with limited computation power.
 
\item A new mapping module to exploit map prediction. We integrate OPNet into previous mapping module by mantaining a doulbe-layered map.  In our experiments, planning with occupancy predictor outperforms both aggressive and conservative planning.
\end{enumerate}

\section{Related Work}
\label{sec:related_work}

\subsection{Planning with Map Uncertainty}
 To guarantee speed and safety simultaneously in unknown cluttered environments, some works have been proposed from the planning perspective in recent years.
Tordesillas et al.~\cite{tordesillas2019faster} propose a method to generate an anaggressive (optimistic) trajectory and an conservative (pessimistic) trajectory restricted within safe space  as a backup  at the same time.
This method costs extra calculations for backup trajectories that are rarely executed.
Some works ~\cite{zhou2020raptor, Oleynikova2018Safe} plan an informative trajectory considering the vehicle's field of view (FoV) to actively observe and avoid unknown obstacles. 
However, these works require an additional mechanism to conduct visibility planning and are indeed conservative planning ways, thus they are still constrained by the sensing range of the vehicles.

In~\cite{richter2018bayesian}, a learning-based method is proposed to predict the risk in the next planning horizon by detecting the novelty of surrounding environments.
Then this estimated risk is used to guide the driving speed of a ground vehicle. 
This method is limited to 2D environments and can only provide limited information, such as the risk or cost, to help the planner make high-level decisions on speed.
Above all, although aiming at the same goal, our work does not conflict with these techniques as our improvement comes from mapping rather than planning.

\subsection{Navigation in Predicted Maps}

 Map prediction shows great potential in the field of exploration. Rakesh et al.\cite{shrestha2019learned} use a predicted map to compute flood-fill information gain to guide exploration. Manish et al. \cite{saroya2020online} use a CNN model to predict topological features in subterranean tunnel networks. Katyal\cite{katyal2019uncertainty} predict the occupancy map beyond the sensor's FoV and then use estimated uncertainty to guide exploration. 
 Nevertheless, all these works operated only in 2D simulators. Hepp et al.\cite{hepp2018learn} do not directly predict the map but use a deep neural network to estimate the utility of viewpoints. These works proves that deep learning models are capable of inferring features of unobserved regions in complex environments from current imperfect perception. 
 
 Katyal et al.\cite{2018Occupancy} try occupancy map prediction with generative and adversarial models. However, their prediction quality is not satisfactory due to lack of training data. Besides, how to utilize map prediction for navigation is not studied in their work.  For planning,  Amine et al.\cite{elhafsi2020map} use a Conditional Neural Process based network to predict potential upcoming turns in maps. By map prediction, their planner can generate smoother and more efficient trajectories in their test environment: single-path 2D mazes with frequent corners and U-turns. Benefitting from map prediction, frequent frontier selection is no long required. The way they explain the utility of map prediction is very similar to ours. However, their simulation environments are extremely ideal and their training and testing environments are almost the same. Thus generalization ability of their method is questionable. In contrast, our proposed method can work in unseen complex 3D environments and real-world experiments. We also designed the whole module systematically which will be introduced in Sec.\ref{section: sec3.3}.

 Above all, all these methods mentioned above are based on supervised learning which requires high quality data.  For instance, in \cite{2018Occupancy}, the authors only manage to create two maps and collect altogether six trajectories. Moreover, obtaining such kind of data in complex 3D scenes is even harder. Our self-surpervised method can learn from public datasets without requirement of ground-truth. In addition, we can generate multiple training data pairs from single original scene.
\begin{table}[hbt]
\renewcommand\arraystretch{1.5}
\begin{tabular}{|l|c|c|}
\hline
                                                   & Property       & Environment             \\ \hline
Katyal et al.\cite{2018Occupancy} & 2D, created    & 6 trajectories in 2maps \\ \hline
Elhafsi et al.\cite{elhafsi2020map}                                     & 2D, created    & 75 single-path mazes    \\ \hline
Ours                                               & 3D, real-world & 80 \textbf{buildings}           \\ \hline
\end{tabular}
\centering
\captionsetup{font={small}}
\caption{Comparison of source of training data. We train our predictor in a self-supervised manner that enables us to use easily accessible data.}
\vspace{-0.5cm}
\end{table}

\subsection{Shape and Scene Completion}
 
  Using limited observation to predict the occluded space can be considered a variant of scene completion task. Although, current scene completion models can hardly run in real-time and lack considerations for a dynamic process like navigation, they represent the most advanced deep learning techniques in this area.
  
 Completing 3D shapes has been well-studied in geometry processing. Many surface reconstruction methods like Poisson Surface Reconstruction\cite{kazhdan2006poisson, kazhdan2013screened} aim to fit a surface and treat point cloud observations as data points in the optimization process. These methods are capable of filling small holes. One of the first data-driven structured prediction methods is Voxlets\cite{firman2016structured}, which uses a random decision forest to predict unknown voxel in a depth image.
 Recently, various deep learning approaches have been developed for scene completion. 
 Song et al. constructed SUNCG\cite{song2017semantic}, a large-scale dataset of synthetic 3D scenes with dense volumetric annotations. 
 They also present SSCNet, an end-to-end 3D convolutional network that takes a single depth image as input and simultaneously outputs occupancy and semantic labels for all voxels in the camera view frustum. However, it requires complete synthetic data and semantic annotations. Recently, ScanComplete\cite{dai2018scancomplete} and SG-NN\cite{dai2020sg} show great capacity for completion of larger missing regions in larger-scale scans. Althogh, SG-NN is faster than any of the methods mentioned above, in real-time application, its latency is still unacceptable.
 
 As in navigation tasks we normally do not need to reconstruct high-resolution object surface or do segmatic segmentation, we need to design lighter and faster models to meet the need of high speed. In addition, we have to balance prediction range, map resolution, and model latency.

\begin{figure*}[t]
\centering
\includegraphics[width=2.0\columnwidth]{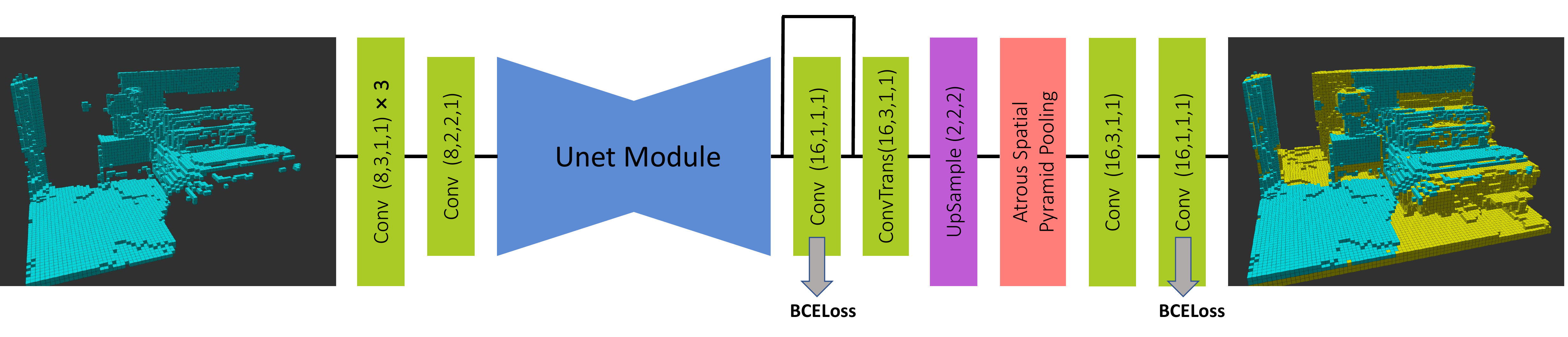}
\captionsetup{font={small}}
\caption{OPNet: Obtacle prediction network. Take a grid map as input and generate a predicted map of the same size. Activation layers and Batch normalization layers are not shown in this figure. The convolution parameters are shown as the number of filters, kernel size, stride, and dilation.
\label{fig:network2}}
\vspace{-0.5cm}
\end{figure*}

\section{Methodology}

In this section we introduce our self-surpervised learning process, OPNet, and the mapping module with online predictor.

In the first part we will introduce the definition of occupancy prediction problem and data generation method.
The second part is about details of the training process and  the proposed OPNet.
After that, we will introduce how to exploit map prediction in navigation, mainly about the design of mapping module.

\subsection{Training Data Generation}

Considering a dense occupancy grid map, $ \mathcal{X} \in \mathbb{R}^3$ represents the spatial
coordinates of the grids in the map, and  $ \mathcal{Y} = \mathrm{\{0,1,-1\}}$ represents their corresponding occupancy. In this paper occupancy value of -1 means \textbf{unknown}. The map $ \mathcal{M} = \{(\mathrm{x}_{i}, \mathrm{y}_{i})\}_{i=1:n}$ consists of occupied space $\mathcal{M}_{obs}\ (\mathrm{y}_{i} = 1)$, free space $\mathcal{M}_{free}\ (\mathrm{y}_{i} = 0)$, and unknown space $\mathcal{M}_{un}\ (\mathrm{y}_{i} = -1)$. 
Defining ground-truth occupancy as $ \mathcal{Y}^{*} = \mathrm{\{0,1\}}$, grids in $\mathcal{M}_{un}$ are indeed free or occupied:
\begin{equation}
\mathcal{M}_{un} = \mathcal{M}_{un}^{+} +  \mathcal{M}_{un}^{-} .\label{def1}
\end{equation}

\begin{equation}
\mathrm{y}_{i}^{*} = \left\{
\begin{array}{rcl}
0, & \mathrm{i} \in \mathcal{M}_{un}^{-},\\
1, & \mathrm{i} \in \mathcal{M}_{un}^{+},\\
\end{array}
\right. \label{def0}
\end{equation}

$\mathcal{M}_{un}^{+}$ is the set of grids that we need to classify. The output of occupancy prediction is:
\begin{equation}
P(\mathrm{y}_{i}^{*} = 1 | \mathcal{M}) \ \mathbf{for} \ \mathrm{i}\ \mathbf{in}\  \mathcal{M}_{un}. \label{def2}
\end{equation}
Taking sensor noise into consideration, the occupancy of all grids may have errors. Thus, the prediction output is:
\begin{equation}
P(\mathrm{y}_{i}^{*} = 1 | \mathcal{M}) \ \mathbf{for} \ \mathrm{i}\ \mathbf{in}\  \mathcal{M}. \label{def3}
\end{equation}

\begin{algorithm}[h]
\caption{Occlusion generation} 
\label{alg:Occlusion}
\hspace*{0.02in} {\bf Input:} 
original map $\widehat{\mathcal{M}}$\\
\hspace*{0.02in} {\bf Output:} 
map with extra occlusion $\widetilde{\mathcal{M}}$
\begin{algorithmic}[1]
\State $ t = 0$, $\widetilde{\mathrm{y}_{i}} = -1 \ \textbf{for} \ i \ \textbf{in} \ \widetilde{\mathcal{M}}$ 
\While {$ t < t_{max}$} 
    \State sample $\mathbf{p}_{start}, \mathbf{p}_{goal}$ in $\widehat{\mathcal{M}}$
  \State find path $\mathcal{T}$ from $\mathbf{p}_{start}$ to $\mathbf{p}_{goal}$
    \If{$collisionCheck(\mathcal{T})$ fail} 
        \State continue
    \Else
\State set $n$ scan points ($\mathbf{p}_{1}, ..., \mathbf{p}_{n}$) on $\mathcal{T}$
\For {i \textbf{in} range($n$)}
        \State $\widetilde{\mathcal{M}}_{i} = simulateObservation(\widehat{\mathcal{M}} ,\mathbf{p}_{i})$
\State $\widetilde{\mathcal{M}} = fuseMap(\widetilde{\mathcal{M}}, \widetilde{\mathcal{M}}_{i})$
\EndFor
\If{ $ r_{min} < knownRatio(\widetilde{\mathcal{M}}) < r_{max}$}
\State Return $\widetilde{\mathcal{M}}$
\Else 
\State Continue
\EndIf
    \EndIf
\EndWhile
\end{algorithmic}
\end{algorithm}

We design the following data generation process to simulate occlusion in a navigation process.
In each scene, a less complete map $\widetilde{\mathcal{M}}$ is generated from the target map$\widehat{\mathcal{M}}$.
Similar to the process in SG-NN\cite{dai2020sg}, we first build room-level TSDF maps with $5cm$ resolution, and then sample $4m \times 4m \times 2m$ blocks in each room as $\widehat{\mathcal{M}}$. 
Please note that $\widehat{\mathcal{M}}$ may contain unobserved space $\widehat{\mathcal{M}}_{un}$, in other words, defects points whose true state is unknown.

\begin{figure}[t]
\centering
\includegraphics[width=0.7\linewidth]{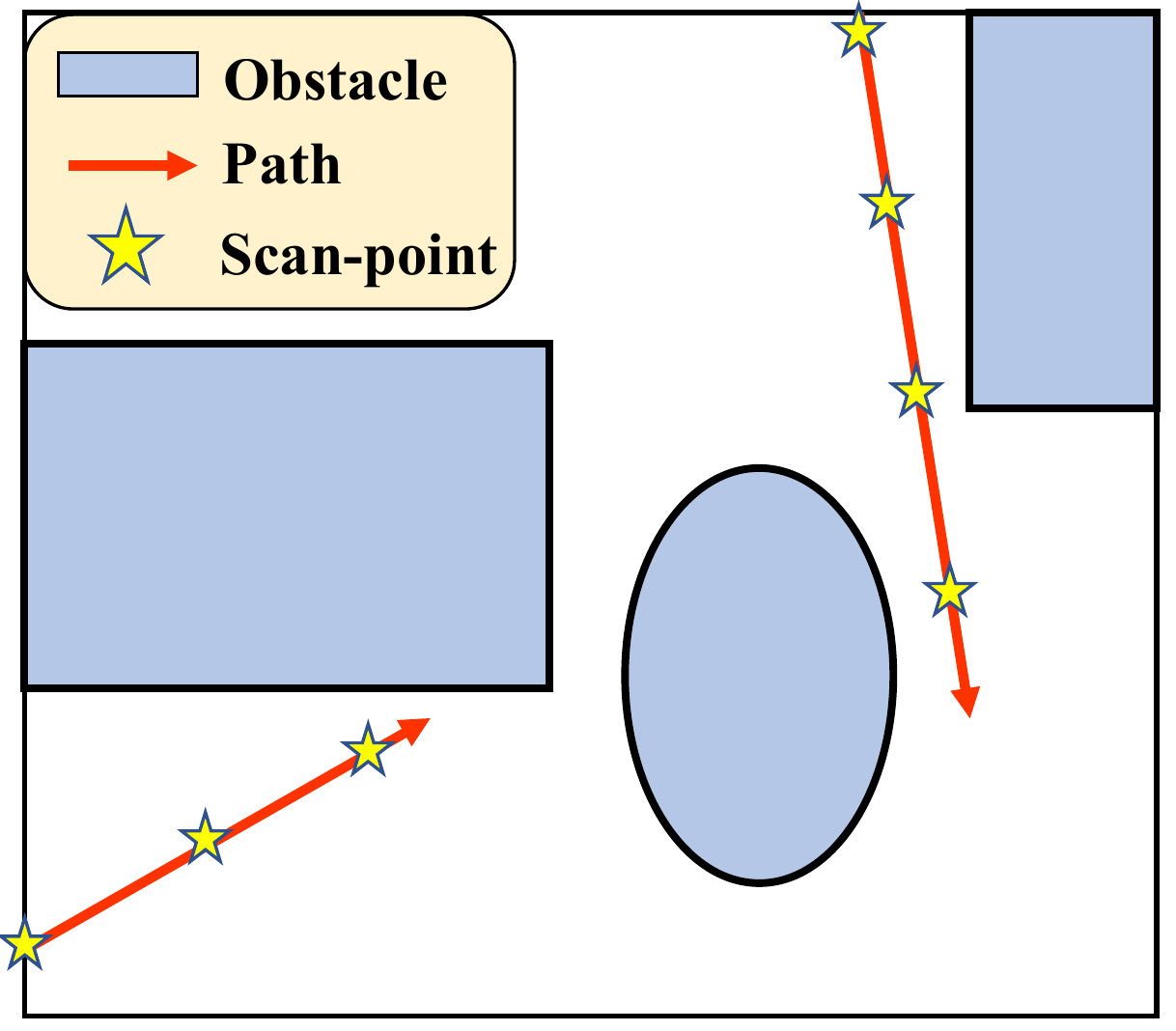}
\captionsetup{font={small}}
\caption{Generating occlusion by simulating navigation process. We sample paths from different dicrections. Then scan-points are set on each path with 1m interval. Scans from scan-points on each path are fused into one less complete map.}
\label{fig:datagen}
\vspace{-0.5cm}
\end{figure}

After that, as shown in Alg.\ref{alg:Occlusion}, occupancy of all grids in $\widetilde{\mathcal{M}}$ is initialized as $-1$.
As shown in Fig.\ref{fig:datagen}, after getting the target map $\widehat{\mathcal{M}}$, we sample a virtual path $\mathcal{T}$ in it and uniformly set a group of scan-points ($\mathbf{p}_{1}, ..., \mathbf{p}_{n}$) with $1m$ interval on it. 
For convience here we use straight path together with a  collision check step. To generate occlusion, we then simulate scans from each of these scan-points considering $\widehat{\mathcal{M}}$ as ground-truth occupancy map and fuse the scans in $\widetilde{\mathcal{M}}$. 
This is done by a reverse raycasting process: unlike standard raycasting process, rays start from $\mathbf{p}_{n}$ towards different directions and stop at either obstacles or unknown grids. 
In this way, some regions in $\widehat{\mathcal{M}}$ become occluded in $\widetilde{\mathcal{M}}$.  
Defining $knownRatio(\widetilde{\mathcal{M}}, \widetilde{\mathcal{M}})$ as the number of known grids in $\widetilde{\mathcal{M}}$ divided by number of known grids in $\widehat{\mathcal{M}}$, we reject  $\widetilde{\mathcal{M}}$ if $knownRatio( \widetilde{\mathcal{M}})$ is lower than $25\%$ or higher than $90\%$ to ensure occlusion quality.

We repeat occlusion generation process for multiple times in each block, randomly sample virtual paths from different directions.

For data generation, we use Matterport3D dataset~\cite{Matterport3D}, an RGB-D dataset containing depth images of building-scale scenes and the corresponding 6-DoF camera poses. 
Because the dataset we use is recorded under very ideal conditions, we add gaussian noise and peper noise on $\widetilde{\mathcal{M}}$ as additional sensor noise.
We generate about 35000 $(\widehat{\mathcal{M}}, \widetilde{\mathcal{M}}) $ pairs from 15,000 blocks using data of 80 buildings in Matterport3D, 65 buildings for the training set and 15 buildings for the validation set.

\subsection{Occupancy Prediction Network}


It is obvious that $\widetilde{\mathcal{M}}_{un} \setminus \widehat{\mathcal{M}}_{un} \neq \emptyset$.
By learning the difference between the input map built by limited scans with massive occlusions and the more complete target map, networks learn to complete the occluded space.

Our predictor network generates occupancy classification for each grid within the map block. 
Its input can be occupancy grids, TSDF, or other kinds of dense tensor, and so is the output. 
For simplicity, in this paper we only introduce occupancy grids, where a value is stored in each grid representing its state. 
All grids in our grid map are initialized as unknown with value -1. 
Every time a new scan comes in, observed grids are updated to a value between 0 and 1: values greater than a threshold means occupied or otherwise free, as general occupancy grid mapping does.
The actual input tensor is discretized into trinary values -1, 0 and 1, representing unknown, free, and occupied.

We use binary cross-entropy (BCE) loss for the predictor's output $\mathcal{M}^{*}$, while TSDF output with smooth ${l1}$ loss has equivalent performance. To focus on completion, we give ``missing" grids that are observed in $\widehat{\mathcal{M}}$ but unobserved in $\widetilde{\mathcal{M}}$ extra loss weight. For occupied grids are relatively sparse in 3D scenes, we also give grids that are occupied in $\widehat{\mathcal{M}}$ extra weight to balance data distribution. Grids that are unobserved in $\widehat{\mathcal{M}}$ do not account for the loss.
\begin{equation}
 \mathcal{L} = \sum_{n=1}^Nw_{n} * BCELoss(\hat{\mathrm{y}_{n}}, \mathrm{y}_{n}^{*}), \label{loss}
\end{equation}
\begin{equation}
 w_{i} = \left\{
 \begin{array}{rr}
0, & \mathrm{i} \in \widehat{\mathcal{M}}_{un},\\
3, & \mathrm{i} \in \widetilde{\mathcal{M}}_{un} \ \setminus \widehat{\mathcal{M}}_{un},\\
3, & \mathrm{i} \in \widehat{\mathcal{M}}_{obs}, \\
1, & otherwise.
\end{array}
\right.  \label{loss-weight}
\end{equation}
We use a U-Net\cite{ronneberger2015u} style architecture, an encoder, and a decoder with skip connections between them. Fig.\ref{fig:network2} shows the general architecture of our model. Rather than lots of convolution layers or large kernels, We use Atrous Spatial Pyramid Pooling (ASPP)\cite{chen2017deeplab} to expand the network’s receptive field and contact contextual information of different scales.
Note that fully convolution networks can take in input of varying sizes at inference time, enabling a trade-off between computational cost and prediction range.
We implement our network in Pytorch and train it on our dataset, using the Adam optimizer and learning rate stepping from $10^{-4}$ to $10^{-3}$. Training for 15 epochs takes around 6 hours on a TITAN X Pascal with a batch size of 20 consuming 7.5 GB of GPU memory.

\subsection{Map Generation and Collision Checking}\label{section: sec3.3}

The map we use for navigation is represented as a double layer dense occupancy grid map. The first layer is the original map $\mathcal{M}_{o}$ fused by raw sensor inputs. The second layer is the predicted map $\mathcal{M}_{p}$ generated from the predictor network which is used for local collision checking during planning.
    
\begin{figure}[t]
\centering
\includegraphics[width=0.8\columnwidth]{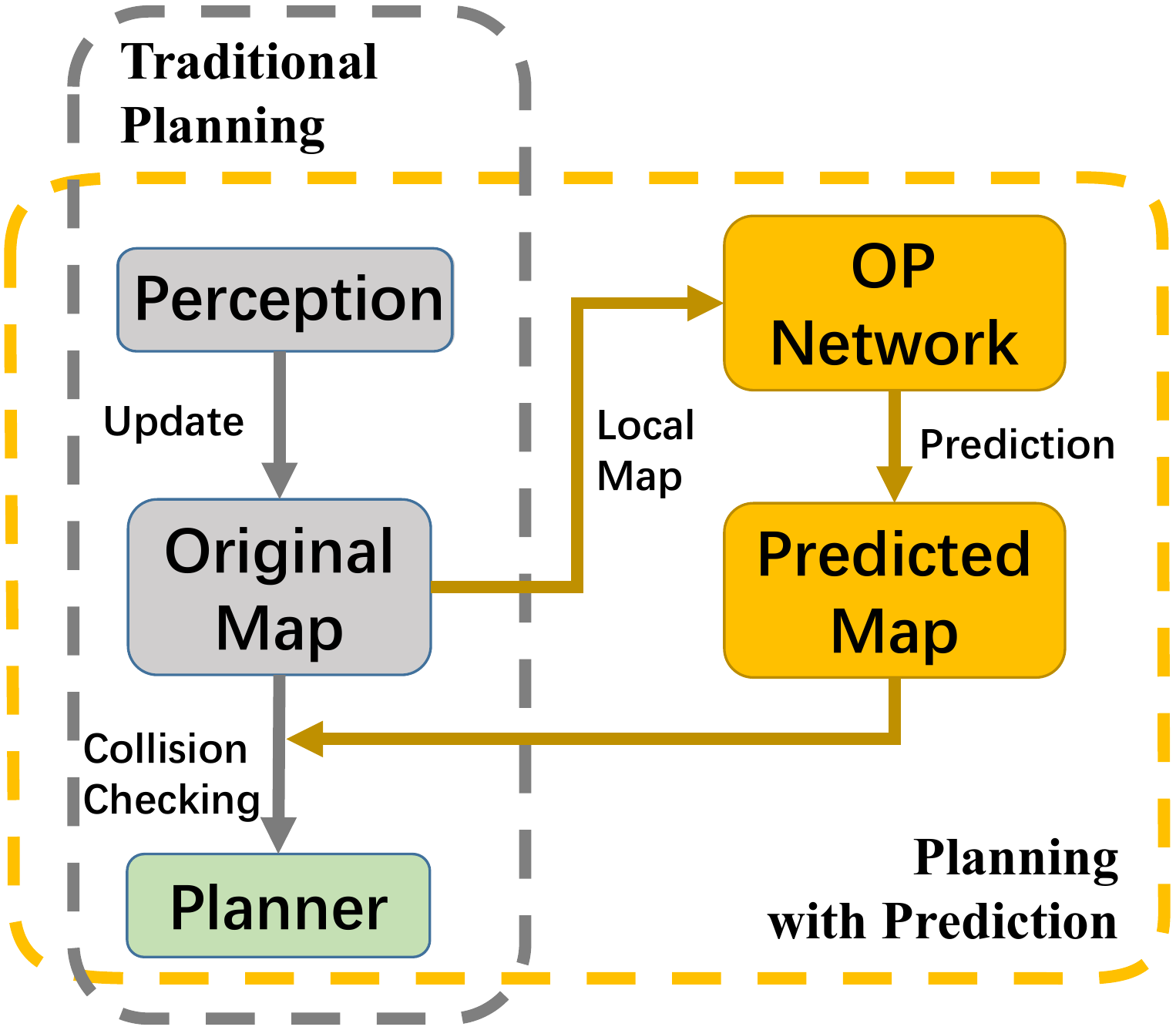}
\captionsetup{font={small}}
\caption{Utilizing map prediction in navigation by maintaining a double layer occupancy grid map. The original map is updated by real sensor perception independently and the predicted map is updated at a fixed frequency on GPU. Collision check fuses values of both layers.\label{fig:system}}
\vspace{-0.8cm}
\end{figure}

All grids in the original layer and the prediction layer are initialized with value -1, meaning unobserved and unpredicted. 
As shown in Fig.\ref{fig:system}, we use asynchronous update: $\mathcal{M}_{o}$ is updated whenever new observation comes in; $\mathcal{M}_{p}$ is triggered at a fixed frequency.
In each prediction loop, a block of voxels consists of trinary values (free, occupied, and unknown) is taken from $\mathcal{M}_{o}$ and fed into the predictor, which generates the probability of being occupied for all grids inside this block.
Then the output of the network is used to update the block of the same position in $\mathcal{M}_{p}$.
The updating frequency is mainly limited by the latency of map prediction and larger prediction rage results in higher latency. 
However, it is natural to adaptively use a larger prediction range with a lower frequency when the robot is moving at a lower speed. 

When performing collision checking in path-finding, both map layers are utilized to embrace richer information of the unknown space and stability of the known space. 
We design the following rules for collision check. 
\begin{enumerate}
\item If a grid is recorded in both layers, a weighted sum of values of both layers is considered: values greater than a threshold means occupied. 
\begin{equation}
y = \lambda_{o}y_{o} +  \lambda_{p}y_{p}, \ \  (y_{o}, y_{p} \neq -1, \lambda_{o} + \lambda_{p} = 1)\label{check0}
\end{equation}
\item If a grid is not observed or predicted, it is considered as free in preferring operating in an optimistic manner. 
\begin{equation}
y = 0, \ \  (y_{o}, y_{p} = -1)\label{check1}
\end{equation}
\item If a grid is only observed or predicted, only the positive value is accepted.
\begin{equation}
y = -y_{p}y_{o}, \ \  (y_{p}y_{o} < 0)\label{check2}
\end{equation}
\end{enumerate}

The weights in (\ref{check0}) can be tuned according to the characteristic of particular sensors. 
For example, if a low error LiDAR is used as we do in our experiment, we raise $\lambda_{o}$ to 0.8 to favor the belief of the observation source. 

It is worth mentioning that, as these two layers are updated asynchronously and separately, delay in the prediction loop will not affect the original map. It means, in the worst case, this double layer map is as complete as a conventional occupancy map with no extra latency.
Above updating and collision check rules can stabilize the frequently changed prediction and provide some expandability.

\section{Benchmark and Experiment}
\label{sec:experiment}
In this section, we compare our model with state-of-the-art scene completion networks in terms of accuracy and inference speed. 
Furthermore, we compare our method with traditional aggressive and conservative planning for UAV and UGV navigation in both simulation and real-world experiments, showing improvement and robustness of our new mapping module.

\begin{figure*}[t]
\centering
\includegraphics[width=0.82\linewidth]{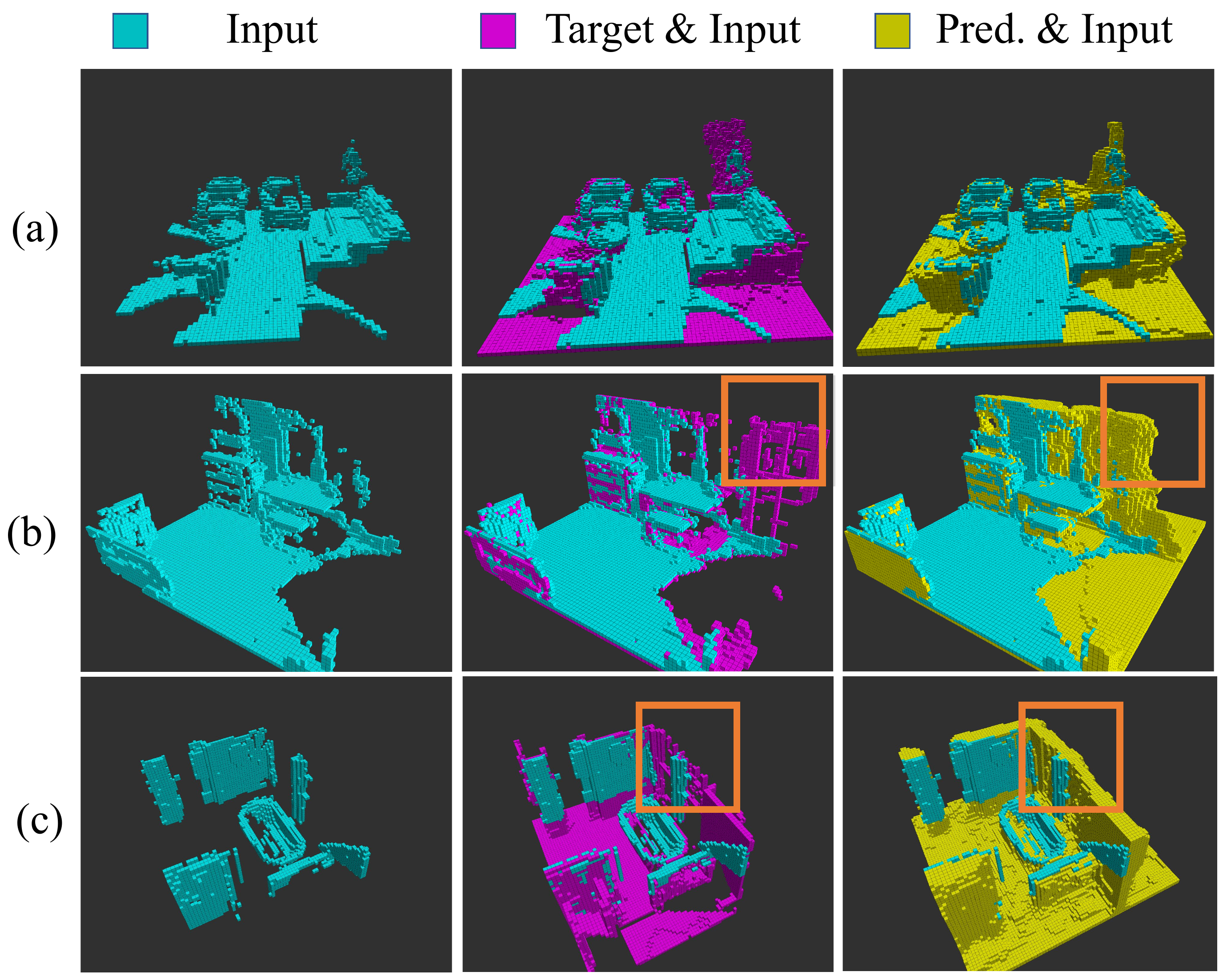}
\captionsetup{font={small}}
\caption{Prediction results in the validation set.
(a) A living room. (b) A desk in front of a window. Part of the upright corner of this image is not successfully predicted because it is totally unobserved. (c) A bathroom. The predicted obstacle is thicker than it should be, and some holes are filled by mistake, which might be glasses of a window.}
\label{fig:pred_result}
\vspace{-0.5cm}
\end{figure*}

\subsection{Prediction of The Unknown}
\begin{table}[b]
\renewcommand\arraystretch{1.5}
\begin{tabular}{|l|c|c|c|c|}
\hline
Method       & prec.(\%)     & recall(\%)    & params        & \begin{tabular}[c]{@{}l@{}}inference \\ time(ms)\end{tabular} \\ \hline
SSCNet       & 64.4          & 46.8          & 231k          & 81.4                                                          \\ \hline
SG-NN        & 54.5          & 31.6          & 216k          & 36.1 $\sim$ 185.8                                              \\ \hline
SG-NN Dense  & 75.6          & \textbf{83.3} & 161k          & 45.6                                                         \\ \hline
SG-NN Dense* & 65.4          & 68.3          & 161k            & 45.6                                                             \\ \hline
OPNet (Ours)         & \textbf{77.6} & 81.6          & \textbf{115k} & \textbf{22.1}                                                 \\ \hline
\end{tabular}
\centering
\captionsetup{font={small}}
\caption{Unknown obstacle prediction results. Input size: 80$\times$80$\times$40. Inferenced on a GTX1650 GPU. 
The only defference between $\textbf{SG-NN Dense*}$  and $\textbf{SG-NN Dense}$ is that its training data is generated for scene completion in the way mentioned in \cite{dai2020sg}.} \label{tabel:network_comparison}
\end{table}

We choose two representative scene completion models, SSCNet\cite{song2017semantic} and SG-NN\cite{dai2020sg} for comparison.
SSCNet is originally trained by complete ground truth and semantic labels on synthetic dataset SUNCG, and SG-NN can be trained on real-world dataset. 
For SSCNet, as semantic segmentation is no longer required, we cut the number of channels of its most convolution layers by half. We use SG-NN's 2-hierarchy levels version rather than the full version with three levels since it is much lighter with almost no performance reduction.
We train these models on our generated dataset, taking occupancy grids as both input and output with a grid size of 0.05 m and block size of 80$\times$80$\times$40. Grids within 0.05 m from the obstacle surface are considered as occupied. 
We also train one model with dataset generated for scene completion to see if it is capable of occupancy prediction.

For evaluation, we only account for missing grids that belong to $\widetilde{\mathcal{M}}_{un} \setminus \widehat{\mathcal{M}}_{un}$. Table.\ref{tabel:network_comparison} summaries the quantitative results and Fig.\ref{fig:pred_result} shows some examples from the validation set. All models are implemented in Pytorch and tested on a GTX1650 GPU. The result of SG-NN is rather odd: although it performs well on scan complete task which it is proposed for, it shows poor capability in our task. We think this is due to the characteristics of sparse convolution. To confirm this conjecture, we replace all sparse convolution operations with dense ones in SG-NN Dense and observe performance as good as ours. 

Results show that our OPNet can reasonably complete partially observed objects. 
In terms of speed, OPNet is at least twice as fast as the fastest variant of SG-NN. Moreover, model trained with our data generation method significantly outperforms previous scene completion model which comfirmed the effectiveness of our occlusion generation method.
Errors mainly occur in the totally unobserved region (Fig.\ref{fig:pred_result}a) or near predicted obstacle surface, for example, thicker or thinner than actual shape (Fig.\ref{fig:pred_result}b).
For real-time usage, our model can run at 20 Hz on an NVIDIA Xavier platform with block size of 80$\times$80$\times$40 and 10 Hz with block size of 120$\times$120$\times$40. 
To be aware, the scale of the dataset that we use for training is relatively small because SUNCG is no longer available anymore. However, our approach is self-supervised, which enables us to use our own sensor scans or synthetic environments to provide extra data.

\subsection{Simulated UAV Navigation}

We conduct simulated 3D UAV navigation in two kinds of scenes: $3m \times 30m$ corridors and a $20m \times 20m$ square rooms, filled with random obstacles which cause massive occlusions.
The UAV needs to fly from one side of the environment to another while avoiding previously unknown obstacles. 
Mapping and planning are conducted on the fly based on instantaneous scans from a simulated laser sensor. 
Re-plan is conducted at a regular frequency, and when the current tracking trajectory is blocked by newly observed obstacles. 
    
    \begin{table}[ht]
\renewcommand\arraystretch{1.3}
    \centering
\begin{tabular}{|l|l|c|c|c|c|}
\hline
\begin{tabular}[c]{@{}l@{}}Environ\\-ment\end{tabular} & Method & \begin{tabular}[c]{@{}l@{}}Succ. \\ Rate \\(\%)\end{tabular} & \begin{tabular}[c]{@{}l@{}}Travel \\ Time \\(s)\end{tabular} & \begin{tabular}[c]{@{}l@{}}Traj. \\ Len. \\ (m)\end{tabular} & \begin{tabular}[c]{@{}l@{}}Emer. \\ Stop \\ Times \end{tabular} \\    
\hline
\multirow{3}{*}{\begin{tabular}[c]{@{}l@{}}Square\\ Room\end{tabular}} & Aggre.   & 86.0 & 11.83  & 30.41 & 1.50 \\ 
\cline{2-6} & Conse. & 70.0 & 20.23 & 33.81 & 4.34 \\ 
\cline{2-6} & Ours & \textbf{92.0} & \textbf{11.79} & \textbf{30.12} & \textbf{0.26} \\ 
\hline \multirow{3}{*}{Corridor} & Aggre. & \textbf{93.0} & \textbf{17.27} & \textbf{28.81} & 0.21\\ 
\cline{2-6} & Conse. & 63.0 & 22.39 & 29.59 & 1.90 \\ 
\cline{2-6} & Ours & 90.0 & 17.50 & 29.00 & \textbf{0.06}\\ 
\hline
\end{tabular}
    \captionsetup{font={small}}
    \caption{Comparison results of simulated UAV navigation experiments.}
\label{tabel:UAV_sim}
\vspace{-0.5cm}
\end{table}

    \begin{figure*}[ht]
\centering
\includegraphics[width=1\linewidth]{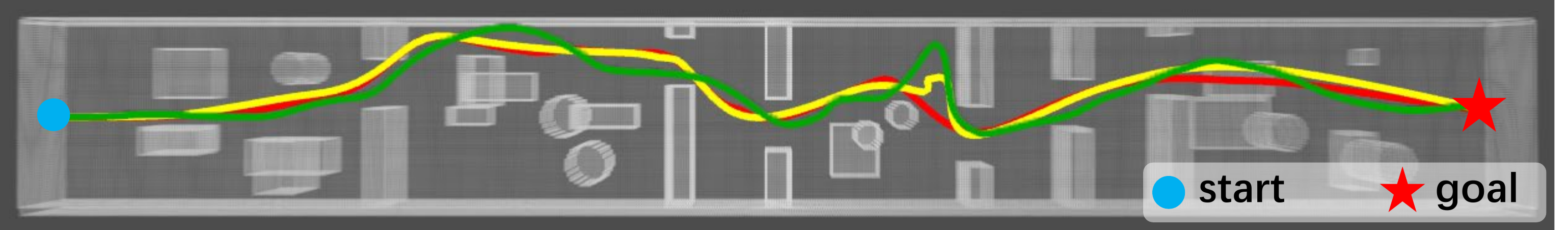}
\captionsetup{font={small}}
\caption{ Some traveled paths in a corridor environment. The proposed method (Red) generates an overall smoother path while both paths generated in the conservative way (Green) and aggressive way (Yellow) are rather winding.}
\label{fig:corridor_planning}
\vspace{0.2cm}
\end{figure*}

\begin{figure}[t]
\centering
\begin{subfigure}{1\linewidth}
\includegraphics[width=1\linewidth]{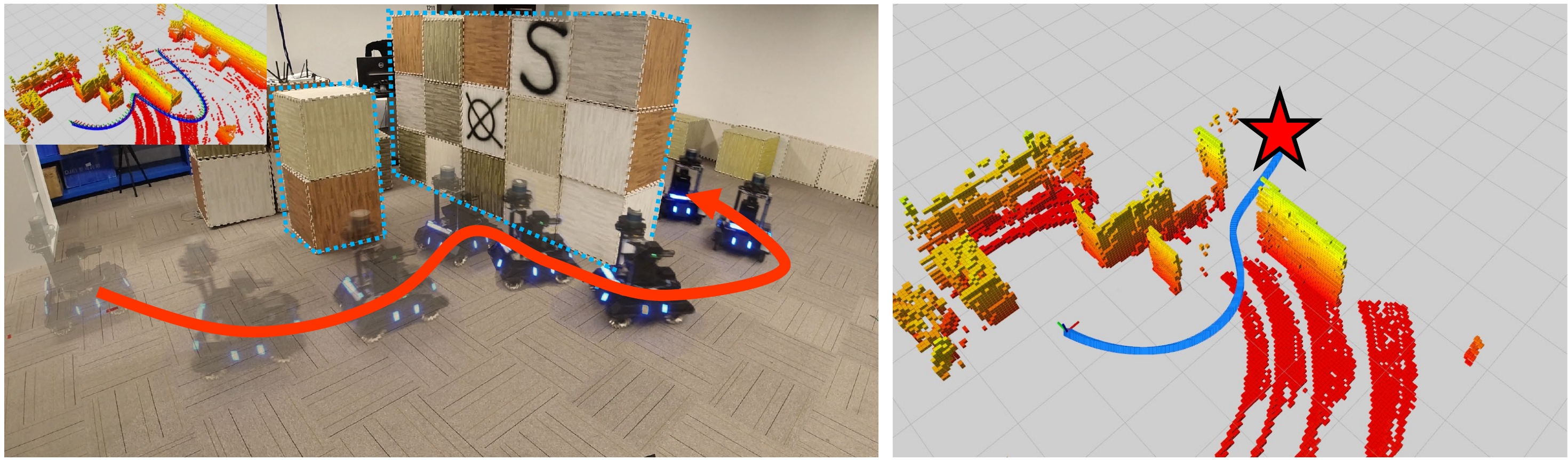}
\caption{Navigate without occupancy prediction in the aggressive manner. The left image shows the winding traveled path. The right image shows the trajectory planed at the beginning crossing the occluded space. Red and yellow grids represent occupancy grids in the original map.}
\vspace{0.3cm}
\label{pic:UGV_wall_jerky}
\end{subfigure}
\begin{subfigure}{1\linewidth}
\includegraphics[width=1\linewidth]{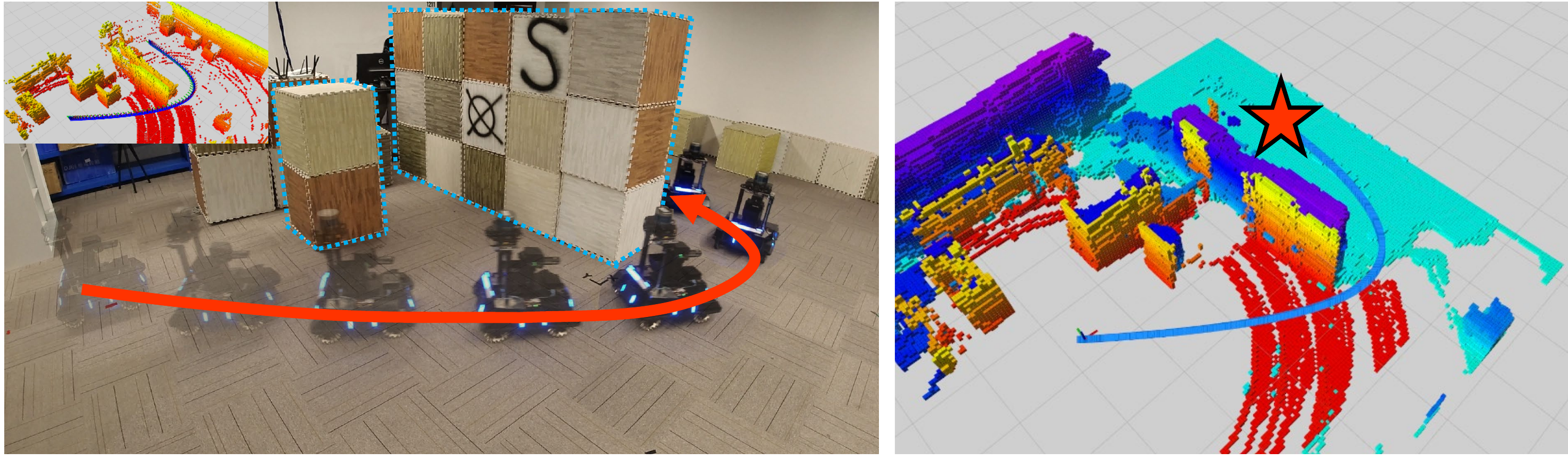}
\caption{Navigate with occupancy prediction. The left image shows the smooth traveled path. The right image shows the trajectory planed at the beginning bypassing the predicted wall. Blue and purple grids represent grids in the predicted map.}
\label{pic:UGV_wall_smooth}
\end{subfigure}
\captionsetup{font={small}}
\caption{Case 1. Navigating to a position behind a wall.}
\label{pic:UGV_wall}
\vspace{-0.2cm}
\end{figure}

\begin{figure}[t]
\centering
\begin{subfigure}{1\linewidth}
\includegraphics[width=1\linewidth]{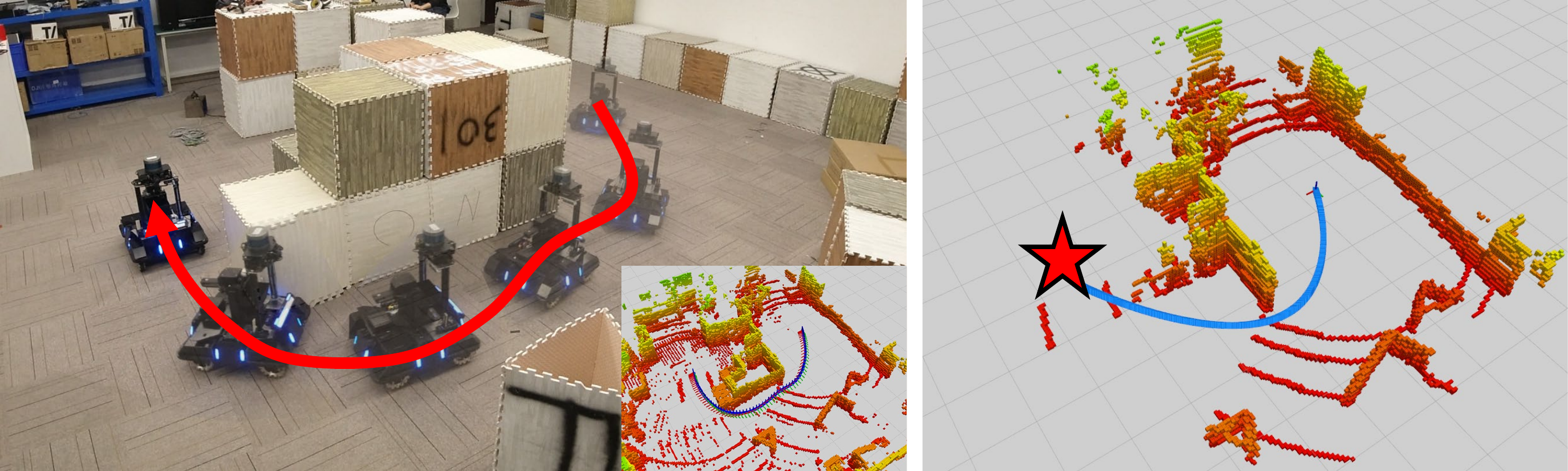}
\caption{Navigate without occpancy prediction. The vehicle plans an initial trajectory deep into the obstacles and only replans after indeed observe the side wall.}
\vspace{0.3cm}
\label{pic:UGV_corridor_jerky}
\end{subfigure}
\begin{subfigure}{1\linewidth}
\includegraphics[width=1\linewidth]{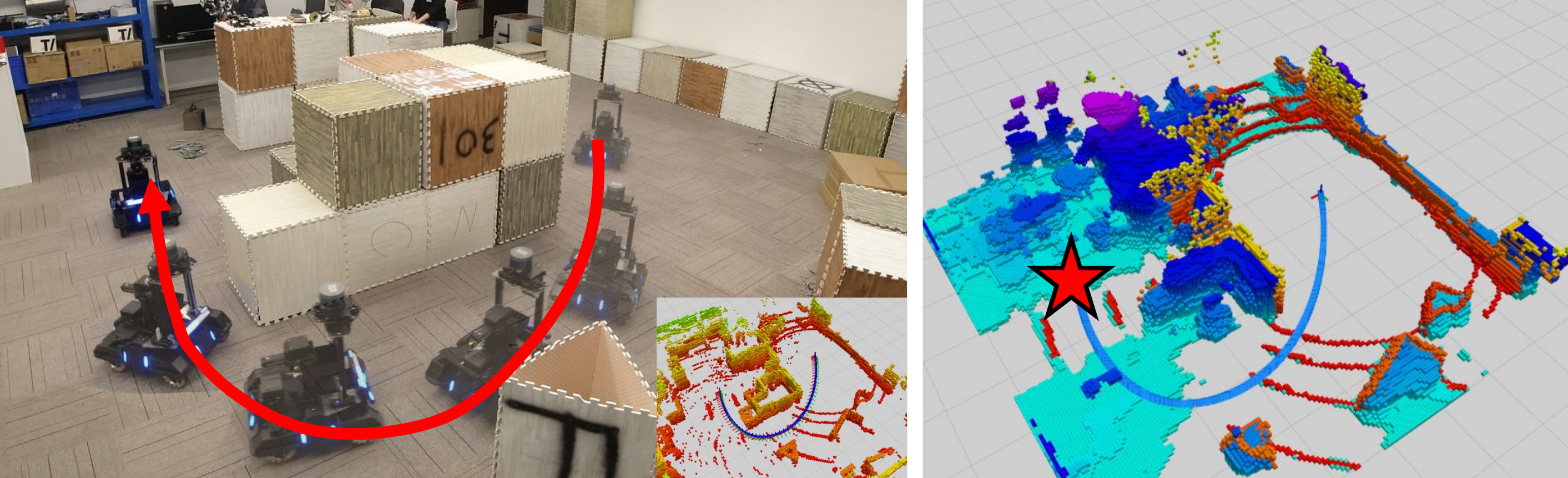}
\caption{Navigate with occupancy prediction. The original occupancy map is represented as the red and yellow grids. The blue and purple grids represent predicted occupied girds. The vehicle avoids the ``unseen" side wall in advance.}
\label{pic:UGV_corridor_smooth}
\end{subfigure}
\captionsetup{font={small}}
\caption{Case 2. Navigating through a corridor.}
\label{pic:UGV_corridor}
\vspace{-0.5cm}
\end{figure}
For collision check, we compare our method with two other schemes that do not have occupancy predicton: the \textbf{ aggressive} way that simply considers unknown space as free, and the \textbf{conservative} way that considers all unknown space as occupied. 
As for the planning module, we adopt a kinodynamic planner~\cite{ye2020tgkplanner} that finds asymptotically optimal trajectories as planning time increases. 
The maximum speed is set as $5m/s$ which can hardly be reached in such cluttered environments.
100 trials with different obstacle placements are conducted in both scenes for each scheme, and we record their average travel time, travel length, number of emergency stops, and success rate for comparison.

Results listed in Table.\ref{tabel:UAV_sim} show that utilizing map prediction in the proposed way, most statistics are improved. Owing to the accurate prediction of the unknown space and avoiding unrevealed obstacles in advance, our method results in fewer emergency stops and a higher success rate compared with the aggressive way while still achieving comparable travel time and overall travel length. Moreover, all aspects outperform the conservative way. The low success rate and high emergency stop times of conservative planning is mainly caused by severe occlusions, leaving strictly limited known space for planning. 
    
An instance of the traveled trajectory is shown in Fig.\ref{fig:corridor_planning}. Plan with the proposed occupancy prediction and mapping module results in overall smoother paths, while aggressive planning usually leads to emergency stops and sharp turns which make its paths winding. Conservative planning, however, influenced by severe occlusions, needs to seek local goals in a small range, leaving its paths winding as well.

\subsection{Real-world UGV Navigation}

UGV experiments are conducted to show that our method is capable of  real-world navigation.
 The UGV platform we use is a DJI Robomaster AI robot\footnote{https://www.robomaster.com/zh-CN/products/components/detail/2499} equipped with a LiDAR\footnote{https://www.robosense.ai/rslidar/rs-lidar-16}, an IMU and a Jetson AGX Xavier\footnote{https://www.nvidia.com/en-us/autonomous-machines/embedded-systems/jetson-agx-xavier/} for all the onboard computing, including localization (done with LIO-SAM~\cite{liosam2020shan}), mapping, network inferring, planning and control. 
 
 Firstly, we find some cases that occupancy predicton improves navigation performance greatly which can vividly explain its benefit. In the scene shown in Fig.\ref{pic:UGV_wall}, the UGV needs to navigate to a place behind the wall. However, its perception is occluded by an obstacle in the front. As a result, it builds an incomplete occupancy map (Fig.\ref{pic:UGV_wall_jerky}) with a ``door" on the wall. 
 The planer consequently plans a shorter trajectory through the ``door", which leads to re-plan afterward, causing a sharp turn that harms localization and control. 
 However, with occupancy predicton (Fig.\ref{pic:UGV_wall_smooth}), the wall is well predicted in advance, resulting in a smooth trajectory that bypasses the wall and avoids unnecessary re-plans.
 In another scene where the UGV needs to bypass a large obstacle shown in Fig.\ref{pic:UGV_corridor}, although only a small part of the side surface is observed in the beginning, the thickness of most of the side surface is predicted (Fig.\ref{pic:UGV_corridor_smooth}). Prediction of the unobserved part of the obscacle facilitates the planner to avoid a premature turn, as will not do without prediction (Fig.\ref{pic:UGV_corridor_jerky}).
 
 Secondly, as shown in Fig.\ref{pic:safe_crash},  we make the UGV operate in a maze-like environment with severe occlusions, making its perception view strictly limited. The UGV successfully travels through the cluttered environment with the help of map prediction providing richer information of the environment. Without prediction, the UGV often goes into a dead end and collide with the wall.
 More details are available in our video.

\section{Conclusion}
In this paper, we propose a self-surpervised learning method to predict occupancy distribution of the unobserved space in obstacle-rich environments. 
Benchmark results show our OPNet achieves higher accuracy and less inference time compared with SOTA models.
We also propose a systematic solution to combine occupancy prediction with the original mapping module to leverage navigation performance. 
The effectiveness of our mapping module is validated by real-time navigation in simulation and real-world experiments. 
    
The main limitation of our network model, however, is the lack of diversity in training data as many deep learning methods do. 
Low similarity between the training dataset and our navigation scenes brings some negative effects to the quality of prediction. 
Nevertheless, our predictor still successfully improved navigation performance without any refinement.
For future work, on one hand, we are looking for more realistic application scenarios like AirSim\footnote{https://github.com/microsoft/AirSim} and real-world  corridors. 
On the other hand, in the future we want to try online learning in challenging environments like forests. 
Moreover, we decide to take the sensors' FoV into account, which should mainly be considered in the data generation step.
We also want to implement our method on a lighter UAV with only depth cameras.

\bibliography{ICRA2021wlz}
\end{document}